\documentclass[conference]{IEEEtran}
\IEEEoverridecommandlockouts

\title{Sample-Efficient Reinforcement Learning Controller for Deep Brain Stimulation in Parkinson's Disease} 
\author{\IEEEauthorblockN{Harsh Ravivarapu\textsuperscript{1}, Gaurav Bagwe\textsuperscript{1}, Xiaoyong Yuan\textsuperscript{1},  Chunxiu Yu\textsuperscript{2*}, Lan Zhang\textsuperscript{1*}}\\
\IEEEauthorblockA{
\textsuperscript{1}
Department of Electrical and Computer Engineering, Clemson University, Clemson, SC;\\
\textsuperscript{2}Department of Biomedical Engineering, Michigan Technological University, Houghton, Michigan;\\
\textsuperscript{*}Corresponding authors: chunxiuy@mtu.edu; lan7@clemson.edu}
}
\date{}
\usepackage[utf8]{inputenc}
\usepackage[T1]{fontenc}

\usepackage{graphicx}
\usepackage{subcaption}
\usepackage[justification=raggedright]{caption}	
\usepackage{nicefrac}
\usepackage{algorithm}
\usepackage{algorithmic}
\usepackage{tabularx}
\usepackage{booktabs}
\usepackage{multirow}
\usepackage{mathtools}
\usepackage{makecell}
\usepackage{enumitem}

\usepackage{bm}
\usepackage{bmpsize}
\usepackage{times}
\usepackage{amsthm}
\usepackage{amssymb,amsmath}

\usepackage{cite}
\usepackage{booktabs}
\usepackage{multirow}

\usepackage{units}
\usepackage{color}

\usepackage{comment}

\usepackage[hyphens]{url}

\usepackage{hyperref}
\usepackage{xspace}

\makeatletter
\DeclareRobustCommand\onedot{\futurelet\@let@token\@onedot}
\def\@onedot{\ifx\@let@token.\else.\null\fi\xspace}

\makeatother
\usepackage{xspace}

\newcommand{\ignore}[1]{}   

\usepackage{bbm}

{\begin{list}               
    {$\bullet$ \hfill}{
        \setlength{\leftmargin}{\parindent}
        \setlength{\parsep}{0.04\baselineskip}
        \setlength{\itemsep}{0.5\parsep}
        \setlength{\labelwidth}{\leftmargin}
        \setlength{\labelsep}{0em}}
    }
{\end{list}}

\providecommand{\cref}[1]{Chapter~\ref{#1}}

\begin{document}

\maketitle
\begin{abstract}
Deep brain stimulation (DBS) is an established intervention for Parkinson’s disease (PD), but conventional open-loop systems lack adaptability, are energy-inefficient due to continuous stimulation, and provide limited personalization to individual neural dynamics.
Adaptive DBS (aDBS) offers a closed-loop alternative, using biomarkers such as beta-band oscillations to dynamically modulate stimulation.
While reinforcement learning (RL) holds promise for personalized aDBS control, existing methods suffer from high sample complexity, unstable exploration in binary action spaces, and limited deployability on resource-constrained hardware.

We propose SEA-DBS, a sample-efficient actor-critic framework that addresses the core challenges of RL-based adaptive neurostimulation. SEA-DBS integrates a predictive reward model to reduce reliance on real-time feedback and employs Gumbel-Softmax-based exploration for stable, differentiable policy updates in binary action spaces. Together, these components improve sample efficiency, exploration robustness, and compatibility with resource-constrained neuromodulatory hardware.
We evaluate SEA-DBS on a biologically realistic simulation of Parkinsonian basal ganglia activity, demonstrating faster convergence, stronger suppression of pathological beta-band power, and resilience to post-training FP16 quantization. Our results show that SEA-DBS offers a practical and effective RL-based aDBS framework for real-time, resource-constrained neuromodulation.
\end{abstract}

\vspace{-1em}

\textit{\textbf{Keywords:
Deep Brain Stimulation, Sample Efficiency, Reinforcement Learning, Gumbel Softmax, Parkinson’s Disease}}

\section{Introduction}
PD is a progressive neurodegenerative disorder characterized by the degeneration of dopaminergic neurons in the substantia nigra, disrupting basal ganglia circuitry and leading to excessive synchronization and elevated beta-band oscillations (12–35 Hz) in cortical-subcortical loops~\cite{brooker2024cell, panicker2021cell, connolly2014pharmacological, mcgregor2019circuit, bouthour2019biomarkers}. These pathological oscillations are strongly correlated with PD’s motor symptoms including bradykinesia, rigidity, and tremor as well as non-motor manifestations such as cognitive impairment, depression, and autonomic dysfunction~\cite{brainsci14070638, okun2010parkinsons}. While the precise etiology remains elusive, PD is widely attributed to a combination of genetic and environmental factors, with increased prevalence among older adults~\cite{brooker2024cell}.

Current therapeutic strategies focus on symptom management. Pharmacological treatments, particularly levodopa-based regimens, remain the standard of care but often lead to diminishing efficacy and motor complications over time~\cite{connolly2014pharmacological}. DBS has emerged as a surgical intervention that delivers continuous high-frequency stimulation to deep brain regions such as the subthalamic nucleus (STN) or globus pallidus internus (GPi). Although effective, conventional DBS operates in an open-loop manner with static parameters, lacking adaptability to fluctuating neural states. This can result in side effects such as dyskinesia, cognitive impairment, and speech disturbances~\cite{kringelbach2007translational}.

To address the limitations of DBS, adaptive DBS (aDBS) systems have been developed to dynamically adjust stimulation parameters based on real-time neurophysiological biomarkers, particularly beta-band power~\cite{little2013adaptive, oehrn2024chronic}. aDBS has shown improvements in therapeutic specificity and energy efficiency~\cite{guidetti2024adaptive, stanslaski2024sensing}; however, most implementations rely on threshold-based or rule-based controllers~\cite{asadi2022origin, radcliffe2023oscillatory, tinkhauser2017modulatory}, which are inherently limited in handling the nonlinear, time-varying nature of brain dynamics.

RL has emerged as a promising paradigm for developing personalized aDBS controllers~\cite{krylov2020reinforcement, Cho2024ClosedLoop}. RL agents can learn closed-loop control policies through interaction with the environment, adapting to individual neural signatures over time. 
However, existing RL-aDBS methods primarily employ model-free algorithms such as Deep Deterministic Policy Gradient (DDPG), which suffer from high sample complexity and limited exploration capabilities. These limitations hinder clinical deployment due to constraints on patient safety, limited interaction budgets, and hardware restrictions~\cite{mehregan2024enhancing}.

To overcome these challenges, we propose \textit{Sample-Efficient Adaptive Deep Brain Stimulation (SEA-DBS)}, a novel RL-aDBS framework designed to accelerate policy learning, enhance beta-band suppression, and enable efficient on-device deployment. SEA-DBS introduces two core innovations: (1) a reward-predictive model that estimates future outcomes from state-action pairs, reducing reliance on direct environment interaction, and (2) a Gumbel-Softmax-based exploration strategy that enables structured, differentiable action sampling in binary control spaces, improving training stability and convergence. We validate SEA-DBS using a biologically realistic simulation of Parkinsonian basal ganglia activity, demonstrating faster convergence, stronger suppression of pathological oscillations, and a significantly reduced memory footprint relative to standard DDPG baselines.

\vspace{1em}
Our primary contributions are as follows:
\begin{itemize}
    \item We address sample inefficiency a key barrier to clinical RL-aDBS by introducing SEA-DBS, a sample-efficient RL framework that integrates predictive reward modeling and Gumbel-Softmax-based exploration for accelerated policy learning.
    
    \item We develop a predictive model to estimate expected rewards from state-action pairs, enabling efficient value updates with reduced reliance on real-world neural stimulation feedback.
    
    \item We incorporate structured, differentiable exploration via Gumbel-Softmax to improve early-stage policy learning in binary stimulation control tasks.
    
    \item We validate SEA-DBS on a biologically realistic simulation of Parkinsonian basal ganglia, demonstrating faster convergence, enhanced suppression of pathological beta-band oscillations, and robustness to FP16 quantization for deployment on resource-limited hardware.
\end{itemize}

The remainder of this paper is organized as follows: Section~\ref{sec:related_work} reviews related work on PD treatment and RL-based neuromodulation. Section~\ref{sec:problem_model} describes the computational PD model and formulates the control problem. Section~\ref{sec:method} presents the SEA-DBS framework, including the predictive model and Gumbel-Softmax strategy. Section~\ref{sec:experiments} details our experimental setup and results. Finally, Section~\ref{sec:conclusion} concludes with a discussion of future directions.

\section{Related Work}\label{sec:related_work}
\noindent \textbf{Stimulation-Based Therapies for PD.} 
As PD progresses, medication alone often becomes insufficient due to motor fluctuations and treatment-resistant symptoms, making stimulation-based therapies a more reliable alternative for sustained symptom management. These approaches can be broadly categorized into:
\textit{1)~conventional DBS}, which delivers continuous high-frequency stimulation to deep brain structures such as the STN or GPi, effectively improving core motor symptoms~\cite{schuepbach2013neurostimulation, chou2013mds}. However, this open-loop system applies fixed stimulation parameters without accounting for dynamic neural states, often resulting in side effects and inefficient energy use~\cite{chen2024framework, gao2023offline}.
\textit{2)~aDBS}, which adapts stimulation in real time using biomarkers such as beta-band activity in local field potentials~\cite{little2013adaptive, oehrn2024chronic}. Most aDBS approaches use threshold-based or rule-based logic to switch or scale stimulation, offering improved efficiency and reduced side effects~\cite{guidetti2024adaptive, stanslaski2024sensing}. Personalized tuning and phase-specific stimulation further enhance control fidelity~\cite{tinkhauser2017modulatory}.

While rule-based aDBS methods offer meaningful improvements over open-loop stimulation, they remain limited by their reliance on hand-crafted heuristics and static control logic motivating learning-based approaches for more adaptive and personalized neuromodulation, such as RL-aDBS.

\noindent \textbf{RL based aDBS for PD.}
RL has recently gained attention in aDBS to optimize stimulation policies based on neural feedback, without relying on fixed thresholds or hand-crafted rules. Unlike traditional aDBS systems that use simple control logic, RL agents can continuously adapt to patient-specific neural dynamics and clinical states, enabling more flexible and personalized neuromodulation strategies~\cite{castano2019simulated}.
Early work by Krylov et al.\cite{krylov2020reinforcement} introduced an open-source RL simulation environment to benchmark DBS control in synthetic PD models. Cho et al.\cite{Cho2024ClosedLoop} demonstrated the use of TD3 in a basal ganglia-thalamic model, achieving effective suppression of pathological oscillations while minimizing energy use. Mehregan et al.~\cite{mehregan2024enhancing} extended this line of work by integrating quantization methods (PTQ, QAT) to enable the deployment of RL-aDBS on resource-constrained neuromodulatory hardware.
While these approaches highlight the potential of RL for improving closed-loop DBS control, they primarily focus on integrating RL for DBS, and do not address challenges like sample efficiency or structured decision boundaries. Addressing these challenges is essential for real-world deployment. In contrast, SEA-DBS proposes a lightweight DDPG-based framework with a reward-predictive model and Gumbel-Softmax exploration to support binary stimulation control and more efficient learning in clinical settings.

\noindent \textbf{Sample-Efficient RL for Clinical Control.}
RL often requires millions of interactions for policy optimization, making it impractical for real-world domains with limited data or costly feedback~\cite{ye2023reinforcement,aastrom2019utilizing}. In clinical neuromodulation, where interactions are sparse, and safety is critical, sample-efficient learning becomes essential.
Several techniques have been proposed to improve sample efficiency in RL. CURL~\cite{laskin2020curl} and SPR~\cite{schwarzer2020data} leverage contrastive and predictive objectives to extract useful representations from high-dimensional observations. Model-based approaches such as SimPLe~\cite{osinski2020model}, Dreamer~\cite{hafner2019dream, hafner2023mastering}, and TD-MPC variants~\cite{hansen2022temporal, hansen2023td} reduce interaction requirements by learning latent dynamics models for planning. EfficientZero~\cite{ye2021mastering} and EfficientZero-v2~\cite{wang2024efficientzero} extend MuZero by integrating consistency losses to enhance generalization under limited data regimes.
In parallel, techniques for handling discrete action spaces such as Gumbel-Softmax~\cite{jang2016categorical},enable differentiable sampling of categorical actions, particularly relevant for binary stimulation decisions in DBS. Recent extensions like GS-FM and GS-SM~\cite{tang2025gumbel} demonstrate structured sequence generation capabilities in domains like DNA and protein design, where sample efficiency and action discreteness are similarly critical.

While these methods show strong performance across synthetic and benchmark environments, their application to clinical RL-aDBS remains limited. Our work adapts predictive learning and Gumbel-Softmax-based sampling to enable efficient policy learning in binary, noisy, and constrained neuromodulatory settings.

\section{PD Brain Model and Problem Statement}\label{sec:problem_model}
\noindent We formulate adaptive DBS as a RL problem, where the agent learns to optimize stimulation based on real-time neural feedback. The computational PD brain model acts as the environment, simulating how neural dynamics evolve in response to stimulation. Biomarkers, specifically beta-band oscillations, define the agent’s observation space, serving as the key input state for decision-making.

\begin{figure}[tb]
    \centering
    \includegraphics[width=0.5\linewidth]{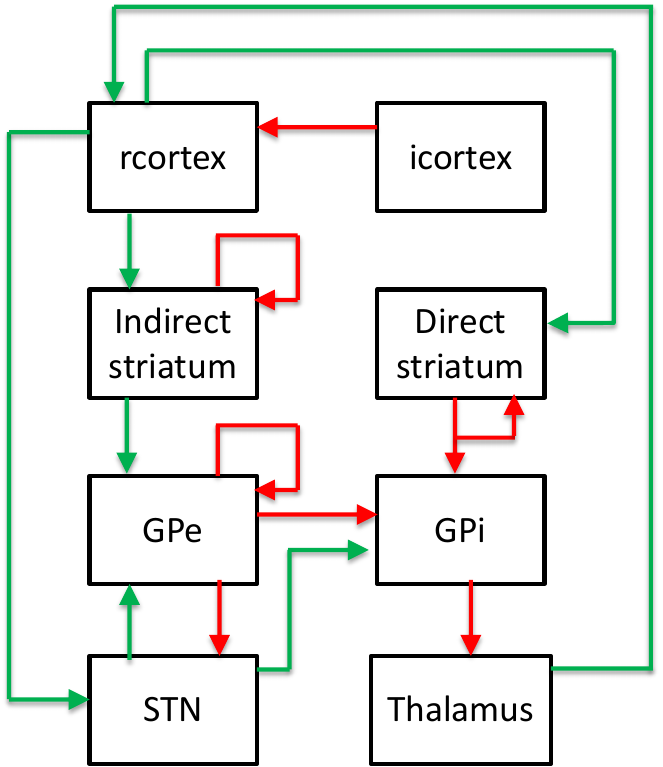}
    \caption{Schematic of the PD brain model, highlighting key excitatory (green) and inhibitory (red) pathways across cortical and basal ganglia regions~\cite{mehregan2024enhancing}.}
    \label{fig:brain model}
    \vspace{-2em}
\end{figure}

\noindent \textbf{Computational Environment - PD Brain Model.}
We adopt a well-established computational model of PD, inspired by Mehregan et al.\cite{mehregan2024enhancing}, that simulates the cortical-basal ganglia-thalamic network under both healthy and Parkinsonian conditions\cite{nagaraj2017seizure, chen2022adaptive, golkhou2004role, golkhou2005neuromuscular}. The model incorporates core brain regions involved in motor control, including excitatory and inhibitory cortical layers, the direct and indirect striatal pathways, the STN, GPi, GPe, and the thalamus. As shown in Figure~\ref{fig:brain model}, each region is modeled using Hodgkin-Huxley-type single-compartment neurons, with ten neurons per population, capturing biologically realistic spiking dynamics.
Both stochastic and deterministic synaptic connections are included to support rich excitatory-inhibitory interactions. DBS is applied to the STN to assess its modulatory effects on pathological beta-band oscillations. Given its ability to reproduce experimentally observed neural activity, this model serves as the RL environment mapping stimulation actions to updated neural dynamics and provides a robust testbed for developing and evaluating RL-aDBS controllers.

\noindent \textbf{RL State from Neural Biomarkers.}
Beta-band oscillations $(13$–$35,\text{Hz})$ in the basal ganglia are widely used as biomarkers for PD, given their strong correlation with motor symptoms such as bradykinesia and rigidity~\cite{chen2010complexity}. Elevated beta activity reflects impaired motor control, while effective DBS reduces beta power, improving movement execution.

In our RL formulation, beta power serves as the observable state input to the agent. To compute it, we analyze the power spectral density (PSD) of spiking activity from GPi neurons. The beta-band power is defined as:
\begin{equation}
P_{\beta} = \frac{1}{n} \sum_{j=1}^{n} \int_{2\pi \cdot 13}^{2\pi \cdot 35} P_{GPi}^{j}(\omega) , d\omega, 
\end{equation} 
where $P_{GPi}^{j}(\omega)$ is the PSD of the $j^{\text{th}}$ GPi neuron, and $n$ is the number of neurons in the GPi region.

\noindent\textbf{Problem Statement.}
Building on the simulated PD environment and beta-band biomarkers, aDBS seeks to modulate stimulation in response to real-time neural dynamics. Despite its promise, practical deployment remains limited due to three key challenges:
\begin{enumerate}[leftmargin=*, labelsep=0.5em]
    \item \textit{Patient-Specific Variability and Biomarker Sensitivity}  
    \noindent Beta oscillations vary significantly across individuals in amplitude, frequency, and phase, and may not consistently reflect all motor symptoms. Stimulation policies must adapt to non-stationary, noisy signals while avoiding overstimulation and conserving energy.
    \item \textit{High Sample Complexity in RL-aDBS}  
    \noindent Standard RL algorithms require frequent interaction with the environment, which is impractical in clinical settings due to patient fatigue and safety concerns. Each stimulation action influences future neural activity, requiring policies that balance short-term efficacy with long-term outcomes.
    \item \textit{Computational and Hardware Constraints}  
    \noindent aDBS devices have strict power and memory limitations, constraining the use of compute-heavy RL algorithms. Many existing methods rely on large replay buffers or batch training, incompatible with real-time, on-device execution.
\end{enumerate}

\section{Methodology}\label{sec:method}
\noindent To address the challenges outlined in Section~\ref{sec:problem_model}, we propose \textbf{SEA-DBS}, an RL framework designed for sample efficient and robust neuromodulation in adaptive DBS. SEA-DBS builds on a DDPG backbone and incorporates two key enhancements: \textit{(i)} a predictive reward model that improves learning efficiency under limited interaction, and \textit{(ii)} Gumbel-Softmax-based exploration for differentiable sampling in discrete action spaces. These components jointly reduce sample complexity, improve exploration, and ensure compatibility with real-time and hardware-constrained clinical settings.
\subsection{Reinforcement Learning Setup for Adaptive DBS}

\noindent We begin by describing our RL formulation in SEA-DBS. RL enables agents to learn stimulation policies through interaction with an environment by maximizing cumulative rewards~\cite{nagaraj2017seizure, chen2022adaptive, golkhou2005neuromuscular}. In the context of aDBS, we model PD treatment as a continuous control task, where the agent adapts stimulation based on real-time neural biomarkers.

We adopt DDPG, an actor-critic algorithm for continuous control problems~\cite{han2020actor, haarnoja2018soft}. The actor-network outputs stimulation decisions (e.g., pulse or no pulse), and the critic network estimates their long-term value using a learned Q-function. These networks are trained through interaction with a simulated PD brain environment, which supplies observations $s_t$, actions $a_t$, and rewards $r_t$ at each timestep. The critic is updated by minimizing the following loss,
\begin{equation}\label{eq:critic_loss}
\mathcal{L}_{\text{critic}} = \text{MSE}\left(Q(s_t, a_t),\ Q_{\text{target}}\right),
\end{equation}
with the target Q-value computed as:
\begin{equation}\label{eq:standard_q_target}
Q_{\text{actor}} = r_t + \gamma Q_{\phi'}(s_{t+1}, \pi_{\theta'}(s_{t+1})),
\end{equation}
where $\gamma$ is the discount factor, and $Q_{\phi'}$, $\pi_{\theta'}$ denote the target critic and actor networks.

To utilize DDPG in the aDBS setting, we define three core components: the state space, action space, and reward function—each grounded in the underlying neural dynamics and clinical constraints.

\noindent \textbf{State Representation.}
The agent observes beta-band power extracted from the GPi region. To capture temporal structure and mitigate short-term variability, we use a fixed-length window of past beta estimates:
\begin{equation}
s_t = \{ P_{\beta}(i) \mid i = 1, 2, \dots, n_{obs} \},
\end{equation}
where $P_{\beta}(i)$ denotes the beta power at timestep $i$. The mean beta power over this window is computed as:
\begin{equation}
\bar{P}_{\beta} = \frac{1}{n_{obs}} \sum_{i=1}^{n_{obs}} P_{\beta}(i),
\end{equation}
and is used as input to both actor and critic networks.

\noindent \textbf{Action Space.}
The agent selects actions from a discrete binary set,
\begin{equation}\label{eq:action}
a_t \in \mathcal{A} = \{0, 1\},
\end{equation}
where $a_t = 1$ corresponds to delivering a DBS pulse and $a_t = 0$ to no stimulation. This action space directly models the binary nature of clinical DBS control, where stimulation is typically applied in short bursts rather than continuously. Each action determines whether a stimulation pulse is administered at the current timestep, allowing the agent to learn temporally sparse stimulation patterns that minimize energy usage while effectively modulating pathological neural activity.

\noindent \textbf{Reward Function.}
The reward encourages the agent to suppress elevated beta activity while avoiding unnecessary stimulation. It is defined as:
\begin{equation}\label{eq:rewards}
r_t =
\begin{cases} 
((\bar{P}_{\beta} - \beta_t) \times 10)^2, & \text{if } \bar{P}_{\beta} < \beta_t \\
-\left((\bar{P}_{\beta} - \beta_t) \times 10\right)^2, & \text{otherwise},
\end{cases}
\end{equation}
where $\beta_t$ is a fixed threshold (set to 0.35). This quadratic formulation penalizes excess beta activity while promoting efficient stimulation.
\begin{figure}[tb]
    \centering
    \includegraphics[width=0.80\linewidth]{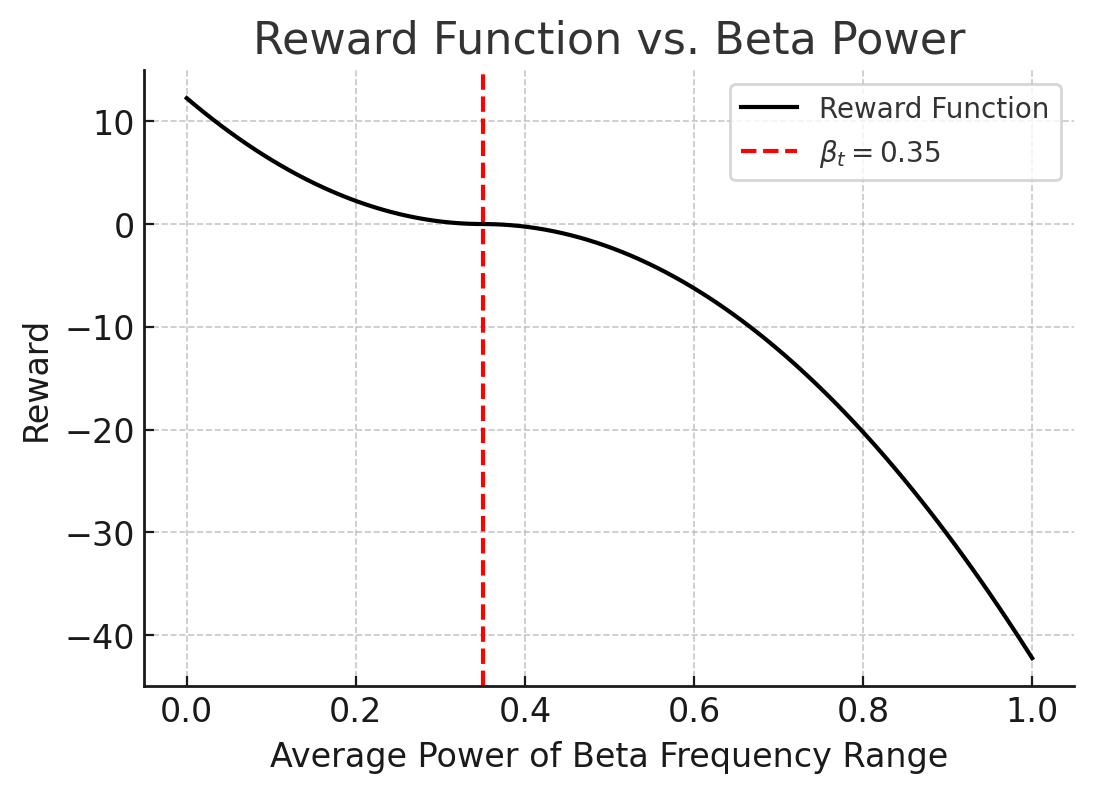}
    \caption{Reward as a function of average beta power $\bar{P}_{\beta}$.}
    \label{reward_poicy}
\vspace{-1.5em}  \end{figure}
Although DDPG provides a structured approach to learning stimulation policies, its reliance on direct reward feedback can limit learning efficiency. To address this, we augment the framework with a predictive model that estimates future rewards, enabling faster policy updates with fewer interactions.

\subsection{Predictive Modeling for Sample Efficiency}
\noindent Predictive modeling estimates future outcomes from past data using learned functions. In RL, predictive models approximate environment feedback such as rewards or state transitions based on state-action pairs. These estimates provide internal signals that can supplement sparse real-world feedback.
To mitigate the dependence on frequent environment interactions, we incorporate a predictive reward model $f_\theta$ that learns to approximate the reward function from observed experience. Given a state-action pair $(s_t, a_t)$, the model generates an estimated reward $\hat{r}_t$ that reflects the agent's expected outcome, enabling additional supervision without requiring direct stimulation feedback, 
\begin{equation}
\hat{r}_t = f_\theta(s_t, a_t).
\label{eq:pred_model}
\end{equation}

We integrate this predicted reward into the Q-target used to update the critic network, modifying the standard formulation to,
\begin{equation}
Q_{\text{target}} = r_t + \hat{r}_t + \gamma Q_{\phi'}(s_{t+1}, \pi_{\theta'}(s_{t+1})),
\label{eq:q_target_pred}
\end{equation}
where $Q_{\phi'}$ and $\pi_{\theta'}$ are the target critic and actor networks updated using actor in Eq. \eqref{eq:standard_q_target}  and critic loss in Eq. \eqref{eq:critic_loss} respectively, and $\gamma$ is the discount factor. Additionally, $r_t$ is the rewards defined using Eq.~\eqref{eq:rewards}.

The predictive model is trained concurrently by minimizing the mean squared error (MSE) between predicted and actual rewards:
\begin{equation}
\mathcal{L}_{\text{pred}} = \frac{1}{B} \sum_{i=1}^{B} (r_i - \hat{r}_i)^2.
\label{eq:pred_loss}
\end{equation}

This auxiliary reward stream supplements sparse or delayed feedback, allowing the agent to learn more efficiently with fewer real-world interactions. As demonstrated in our ablation analysis (Section~\ref{sec:experiments}), predictive modeling substantially accelerates policy convergence and improves beta-band suppression, especially during early training stages with limited data.

\subsection{Structured Exploration for Efficient Policy Optimization}

\begin{figure}[tb]
    \centering\includegraphics[width=0.85\linewidth]{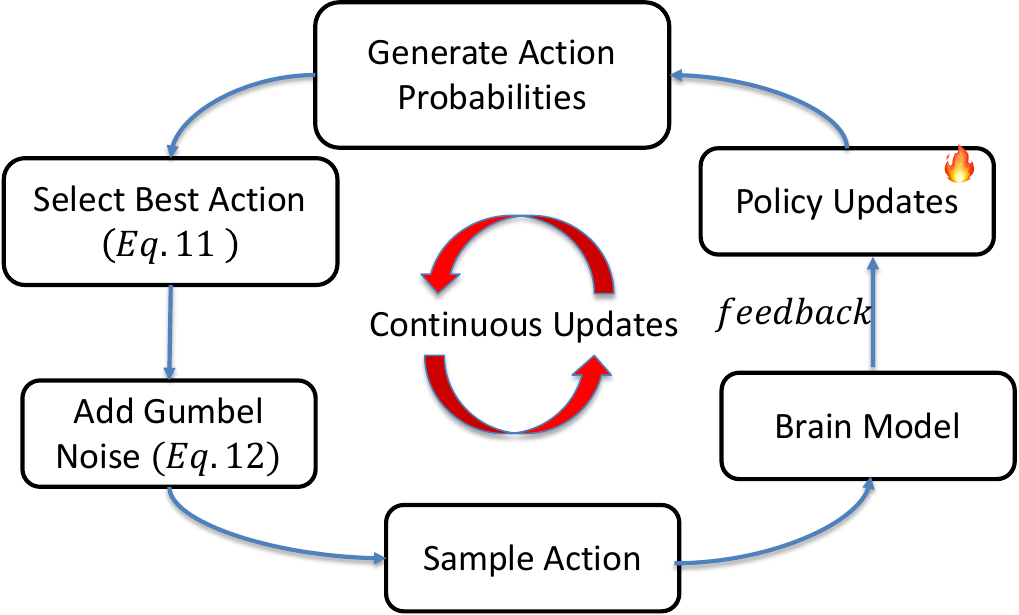}
    \caption{Overall training process using predictive modeling and Gumbel-Softmax: (1) Action probabilities are generated using the states from the brain model, (2) Gumbel noise is added to the action probabilities, (3) Actions are sampled from the resulting distribution, executed by the brain model, and feedback is used to iteratively update the policy.
}
    \label{fig:gumbel-softmax}
    \vspace{-1.75em}
\end{figure}

\noindent
Exploration is a central challenge in RL, particularly during early training when the agent must evaluate diverse action trajectories to avoid suboptimal convergence. This is especially critical in low-interaction regimes such as aDBS, where inefficient exploration leads to poor generalization and unstable policies.
To support gradient-based learning in such settings, we adopt the Gumbel-Softmax distribution, which provides a differentiable approximation of categorical sampling. GS enables stochastic action selection while allowing gradients to flow through non-deterministic choices. Given class probabilities $\pi_1, \cdots, \pi_k$, a relaxed sample $\tilde{a}_i$ is computed as, 
\begin{equation}
\tilde{a}_i = \frac{\exp\left((\log \pi_i + g_i)/\tau\right)}{\sum_{j=1}^{k} \exp\left((\log \pi_j + g_j)/\tau\right)},
\label{eq:gumbel_softmax}
\end{equation}
where $g_i$ is Gumbel noise sampled as:
\begin{equation}
g_i = -\log(-\log(U_i)), \quad U_i \sim \text{Uniform}(0,1),
\label{eq:gumbel_noise}
\end{equation}
where $\tau > 0$ is a temperature parameter controlling the output smoothness. As $\tau \rightarrow 0$, the output approaches a discrete one-hot vector; larger $\tau$ values produce softer distributions.

In RL, GS is particularly useful for discrete action spaces, where standard gradient-based methods cannot directly optimize non-differentiable sampling. This is especially relevant in aDBS, where the action space is binary (stimulate or not). Conventional methods such as DDPG are designed for continuous control and struggle with effective exploration in such discrete domains, often leading to suboptimal convergence.

To enable efficient exploration while preserving end-to-end differentiability, we integrate GS into the policy network. Specifically, the actor outputs logits $\log \pi_\theta(s)$, which are perturbed with Gumbel noise and transformed via temperature-scaled softmax,
\begin{equation}
z_i = \frac{\log(\pi_\theta(s_i)) + g_i}{\tau}.
\label{eq:gumbel_logits}
\end{equation}
These $z_i$ values are then passed through Eq.~\ref{eq:gumbel_softmax} to produce stochastic yet differentiable action probabilities.

To balance exploration and exploitation during training, we anneal the temperature $\tau$ over time:
\begin{equation}
\tau_t = \max(\tau_{\min}, \tau_0 e^{-\lambda_{\tau} t}),
\label{eq:annealing}
\end{equation}
where $\tau_{\min}$ mitigates premature convergence and $\lambda_{\tau}$ governs the annealing schedule. Initially, a high $\tau$ promotes exploration; over time, decreasing $\tau$ sharpens the policy distribution, facilitating stable exploitation.

\subsection{SEA-DBS Overall Framework}
\begin{algorithm}[tb]
    \caption{Training Procedure for SEA-DBS}
    \label{alg:edbs}
    \begin{algorithmic}[1]
        \STATE \textbf{Initialize:} Actor network $\pi_{\theta}$, Critic network $Q_{\phi}$, Predictive model $f_{\theta}$, Replay buffer $\mathcal{D}$.
        \FOR{each episode $e = 1$ to $N_{\text{episodes}}$}
            \STATE Reset environment and obtain initial STATE $s_0$.
            \FOR{each step $t = 1$ to $T$}
                \STATE Compute annealed temperature: Eq. \eqref{eq:annealing}
                \STATE Obtain action logits: $a_{\text{logits}} = \pi_{\theta}(s_t)$.
                \STATE Sample action using Gumbel-Softmax: \eqref{eq:gumbel_softmax} 
                \STATE Execute action $a_t$, observe next STATE $s_{t+1}$ and reward $r_t$.
                \STATE Predict future reward using the predictive model: Eq. \eqref{eq:pred_model}
                \STATE Store transition $(s_t, a_t, a_{\text{logits}}, r_t, \hat{r}_t, s_{t+1})$ in replay buffer $\mathcal{D}$.
                \STATE \textbf{Update Networks:}
                \STATE Sample mini-batch from $\mathcal{D}$.
                \STATE Compute target Q-value: Eq. \eqref{eq:q_target_pred}
                \STATE Compute critic loss: $L_{\text{critic}} = \frac{1}{B} \sum_{i=1}^{B} (Q_{\phi}(s_i, a_i) - Q_{\text{target}})^2$
                \STATE Compute actor loss: $L_{\text{actor}} = - \mathbb{E}_{s \sim \mathcal{D}} [ Q_{\phi}(s, \pi_{\theta}(s)) ]$
                \STATE Compute predictive model: Eq. \eqref{eq:pred_loss}          
                \STATE Update actor, critic, and predictive model using gradients.
                \STATE Soft update target networks: 
                \STATE \hspace{1em} $\theta' \gets \tau \theta + (1 - \tau) \theta'$
                \STATE \hspace{1em} $\phi' \gets \tau \phi + (1 - \tau) \phi'$
            \ENDFOR
        \ENDFOR
    \end{algorithmic}
    
\end{algorithm}

\noindent
SEA-DBS enhances a standard DDPG backbone by incorporating predictive modeling and structured exploration to enable sample-efficient and stable policy learning for aDBS. The framework jointly addresses the high cost of real-world interactions and exploration in low-data regimes.

The overall training pipeline is described in Algorithm~\ref{alg:edbs}.
At the beginning of each episode (Line~2), the environment is reset and the initial state $s_0$ is observed. For each timestep, the agent first computes the annealed temperature $\tau_t$ based on Eq.~\eqref{eq:annealing} (Line~5), which controls the smoothness of the exploration strategy. The actor network $\pi_\theta$ generates action logits from the current state $s_t$ (Line~6), and a differentiable action is sampled using Gumbel-Softmax (Line~7; Eq.~\eqref{eq:gumbel_softmax} and Eq.~\eqref{eq:gumbel_logits}). The selected action $a_t$ is then applied to the environment, which returns the next state $s_{t+1}$ and observed reward $r_t$ (Line~8).

To reduce reliance on real-time feedback, the predictive model $f_\theta$ estimates an auxiliary reward $\hat{r}_t$ using Eq.~\eqref{eq:pred_model} (Line~9). The transition $(s_t, a_t, a_{\text{logits}}, r_t, \hat{r}_t, s_{t+1})$ is stored in the replay buffer $\mathcal{D}$ (Line~10) for later training.
During the update phase (Lines~11–19), a mini-batch is sampled from $\mathcal{D}$ (Line~12). The Q-target is computed using both the observed and predicted rewards via Eq.~\eqref{eq:q_target_pred} (Line~13). The critic is trained to minimize the Bellman error (Line~14; Eq.~\eqref{eq:critic_loss}), while the actor is optimized by maximizing the critic’s evaluation of its actions (Line~15). The predictive model is updated to minimize the prediction error in Eq.~\eqref{eq:pred_loss} (Line~16). All networks are updated using gradients, followed by soft target updates (Lines~17–19).

By integrating synthetic reward estimation and differentiable action sampling, SEA-DBS reduces sample complexity and improves policy robustness under clinically constrained interaction budgets. This makes it suitable for real-time closed-loop neuromodulation where data efficiency and safety are critical.

\section{Experiments}\label{sec:experiments}
\subsection{Experiment Setup}

\noindent \textbf{RL Environment.}
To evaluate the effectiveness of SEA-DBS, we employ the biophysical model mentioned in Section~\ref{sec:problem_model}. The basal ganglia consist of 10 neurons, with stimulation applied to the STN. The environment is designed to mimic PD patients' oscillatory activity, allowing the RL agent to learn stimulation patterns that suppress pathological beta-band oscillations.
The agent observes neural dynamics through the average beta-band power computed from the GPi population. At each timestep, the state $s_t$ consists of a fixed-length window of past beta power values, processed using PSD estimation. This representation provides temporal context while smoothing short-term variability.

\noindent \textbf{Action and Reward.}
The agent selects binary stimulation actions $a_t \in \{0, 1\}$, where 1 denotes delivering a DBS pulse and 0 corresponds to no stimulation. The reward function (Eq.~\eqref{eq:rewards}) encourages the suppression of average beta power $\bar{P}_\beta$ below a clinical threshold $\beta_t = 0.35$, penalizing overstimulation while promoting energy efficiency.

\noindent \textbf{Training Procedure.}
Each episode simulates $60~\text{ms}$ of brain activity, divided into 30 environment steps of $2~\text{ms}$ each. Within each step, the underlying neural dynamics are integrated at a resolution of $0.02~\text{ms}$ to accurately capture high-frequency fluctuations.

The agent is trained over 150 episodes using the DDPG-based SEA-DBS framework described in Algorithm~\ref{alg:edbs}. The training loop incorporates three core components: predictive reward modeling to enhance sample efficiency, Gumbel-Softmax-based stochastic action sampling for improved exploration, and standard critic-actor updates for policy learning. All networks are initialized randomly and optimized using the hyperparameters listed in Table~\ref{tab:hyperparams}.

\begin{table}[tb]
\centering

\begin{tabular}{lr}
\toprule
\textbf{Hyperparameter} & \textbf{Value} \\ 
\midrule
Actor Learning Rate ($\alpha_{a}$) & 0.0005\\ 
Critic Learning Rate ($\alpha_{c}$) & 0.001\\ 
Discount Factor ($\gamma$) & 0.99 \\ 
Replay Buffer Size & 8192 \\ 
Batch Size & 32 \\ 
Exploration Strategy & Gumbel-Softmax with Annealing\\ \bottomrule
\end{tabular}
\caption{Training Hyperparameters.}
\vspace{-2em} 
\label{tab:hyperparams}
\end{table}

\begin{figure}[tb]
    \centering
    \begin{subfigure}{1\linewidth}
        \centering
        \includegraphics[width=0.80\linewidth]{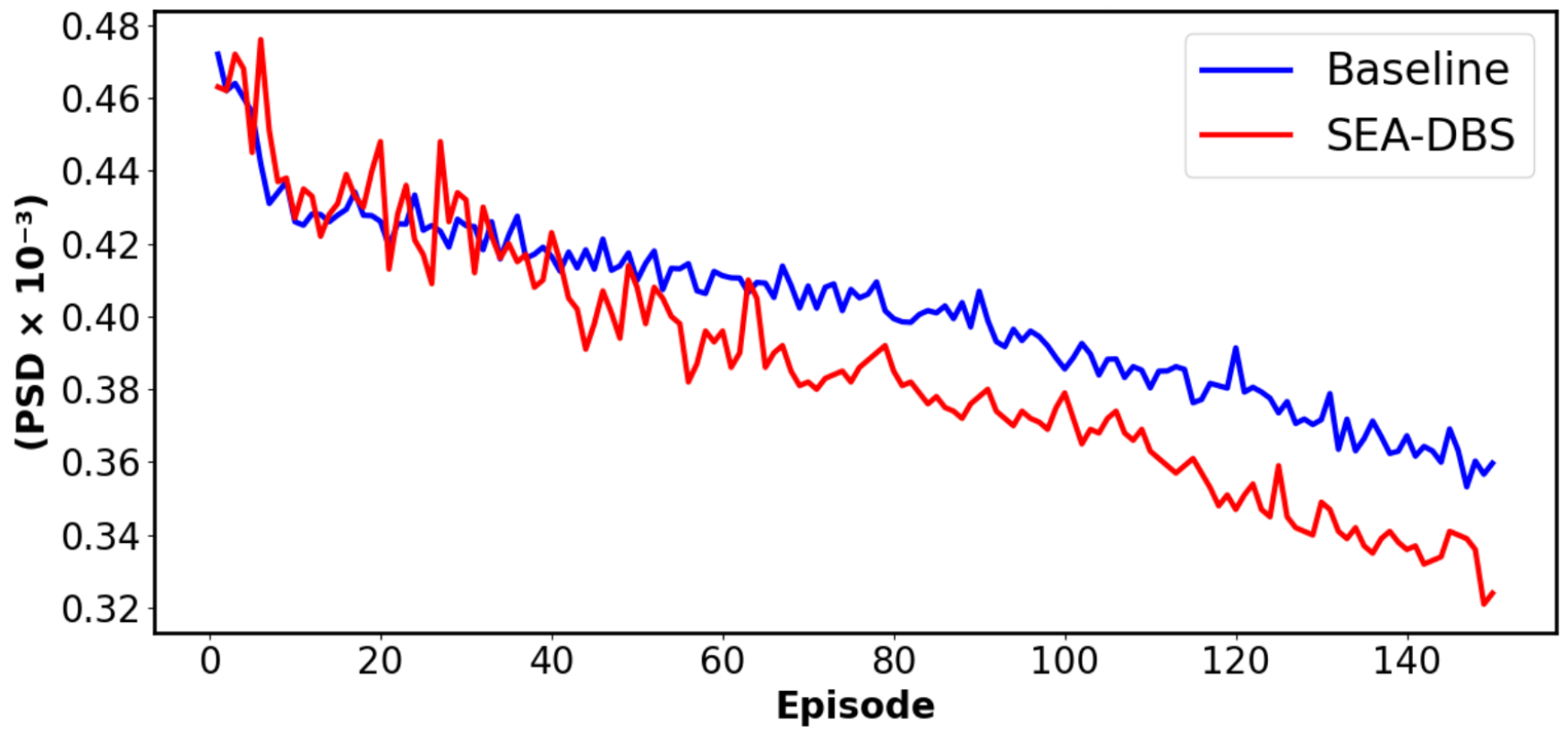}
        \caption{PSD over training episodes.}
        \label{fig:training_psd}
    \end{subfigure}
    \vfill
    \begin{subfigure}{1\linewidth}
        \centering
        \includegraphics[width=0.80\linewidth]{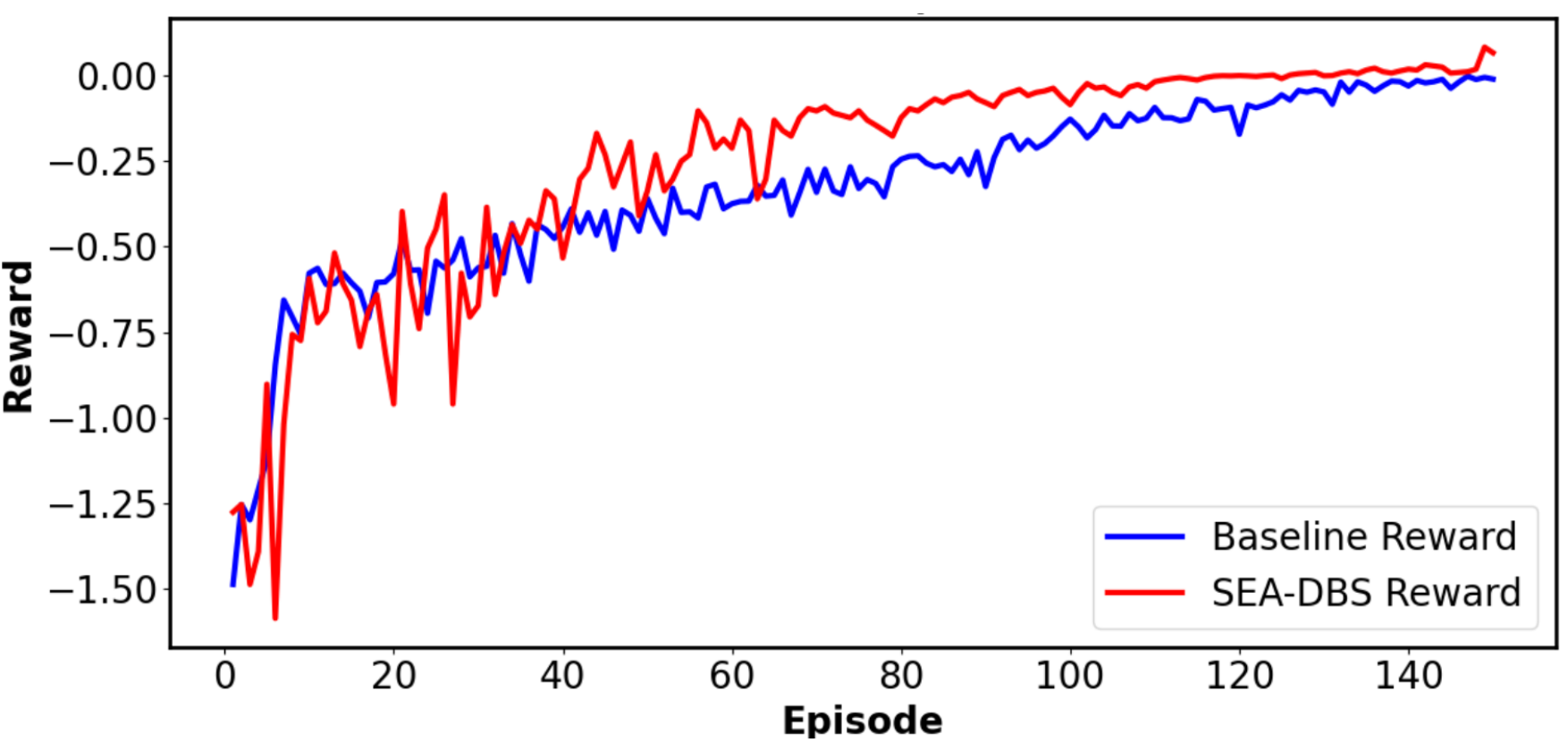}
        \caption{Reward during training.}
        \label{fig:training_reward}
    \end{subfigure}
    
    \caption{Comparison of SEA-DBS model training.}
    \label{fig:training}  \end{figure}

\subsection{Evaluation Results}
\noindent \textbf{Efficiency Analysis.} 
Figure \ref{fig:training}(a) illustrates the average PSD in the beta band across training episodes for both the Baseline and SEA-DBS models. The SEA-DBS model demonstrates a more pronounced and consistent suppression of beta activity over time, indicative of its enhanced ability to disrupt pathological neural synchrony associated with Parkinson’s disease. In contrast, the Baseline model exhibits only a modest decline in beta power. Figure \ref{fig:training}(b) presents the corresponding training reward trends, showing that SEA-DBS achieves higher cumulative rewards with a faster initial learning rate. This suggests that the reinforcement learning agent governing SEA-DBS not only learns more effectively but also adapts stimulation parameters more efficiently to achieve therapeutic goals. Together, these results underscore the efficacy of SEA-DBS in both modulating pathological neural activity and optimizing clinical outcomes through adaptive learning.

\begin{table}[tb]
    \centering
    \begin{tabular}{c@{\hskip 12pt}cc@{\hskip 12pt}cc}
        \toprule
        \multirow{2}{*}{Steps} & \multicolumn{2}{c}{Baseline (DDPG)} & \multicolumn{2}{c}{SEA-DBS (Ours)} \\
        \cmidrule(lr){2-3} \cmidrule(lr){4-5}
        & Avg PSD $\downarrow$  & \begin{tabular}{c} Avg Reward $\uparrow$ \\ $(\times10^{-2})$ \end{tabular} & Avg PSD $\downarrow$ & \begin{tabular}{c} Avg Reward $\uparrow$ \\ $(\times10^{-2})$ \end{tabular} \\
        \midrule
        10  & 364 & -1.96 & \textbf{326} & \textbf{5.76} \\
        20  & 366 & -2.56 & \textbf{309} & \textbf{16.81} \\
        50  & 332 &  3.24 & \textbf{304} & \textbf{21.16} \\
        75  & 328 &  4.84 & \textbf{302} & \textbf{23.00} \\
        \bottomrule
    \end{tabular}
    \caption{
        Performance comparison between Baseline (DDPG) and SEA-DBS across different stimulation update intervals. SEA-DBS achieves lower PSD and higher average reward, indicating more effective and efficient neuromodulation.
    }
    \label{tab:sea_psd_reward}
    \vspace{-1.75em} 
\end{table}

\begin{figure}[tb]
    \centering
    \begin{subfigure}{0.80\linewidth}
        \centering
        \includegraphics[width=\linewidth]{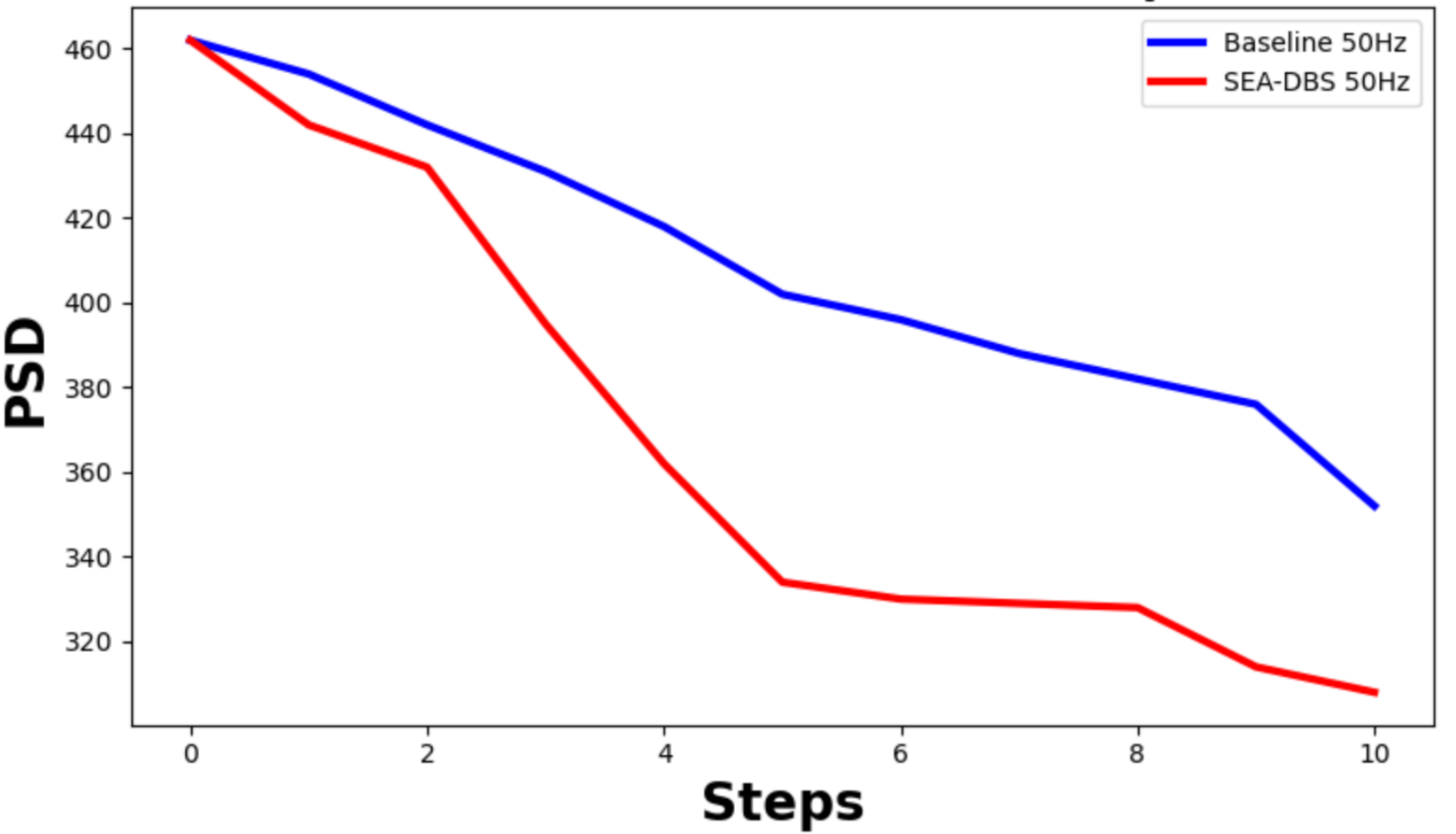}
        \caption{ Beta stimulation freq. 50 Hz.}
        \label{fig:testing_50hz}
    \end{subfigure}
    \hfill
    \begin{subfigure}{0.80\linewidth}
        \centering
        \includegraphics[width=\linewidth]{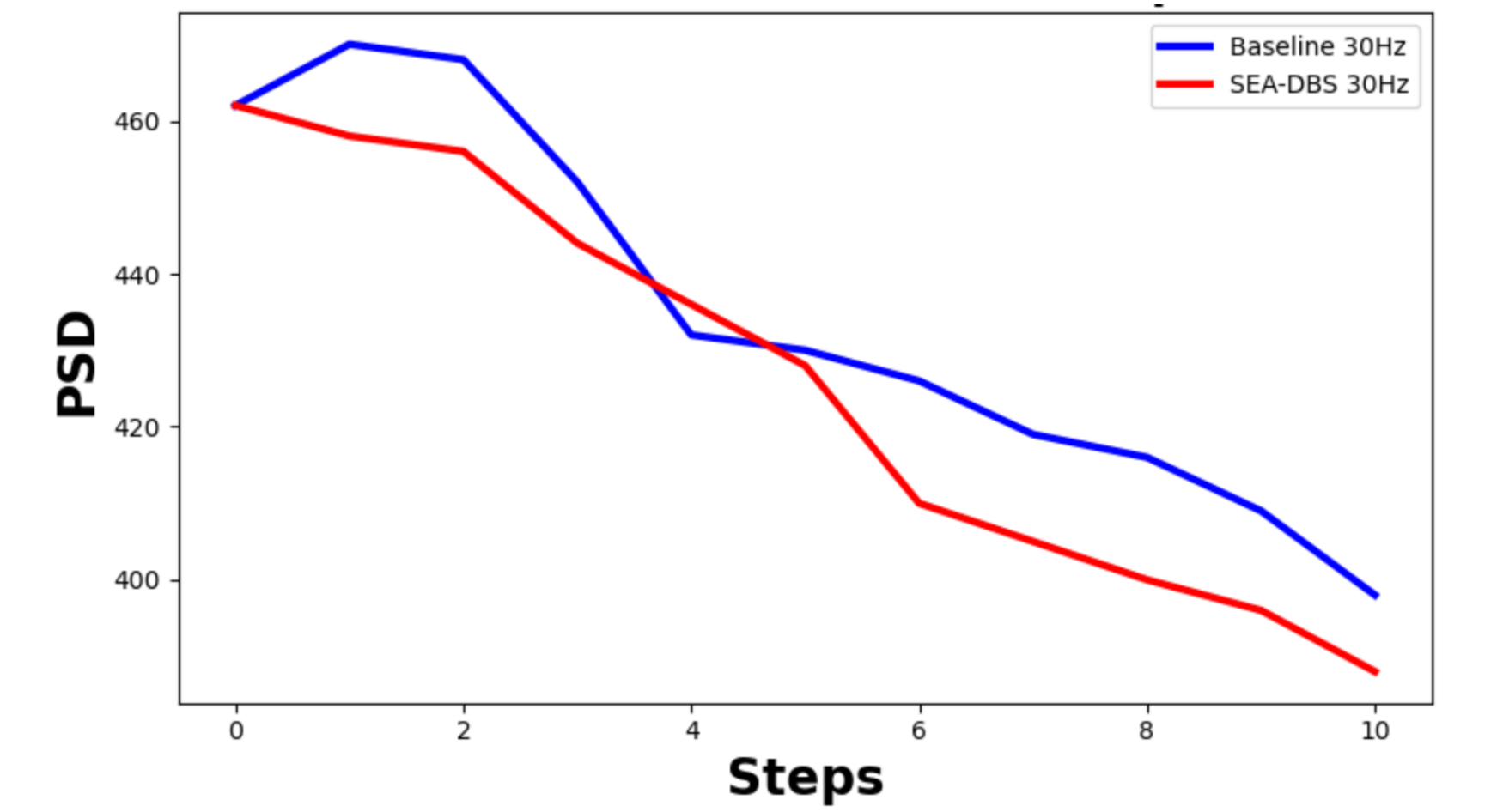}
        \caption{Beta stimulation freq. 30 Hz.}
        \label{fig:testing_30hz}
    \end{subfigure}
    
    \caption{Inference comparison of SEA-DBS vs Baseline.}
    \label{fig:testing}
\vspace{-1.75em}  \end{figure}

We evaluate SEA-DBS to simulate varying PD progression rates by changing the environment seed every $n\in\{10, 20, 50, 75\}$ step. Table~\ref{tab:sea_psd_reward} reports the average PSD and average reward under each setting. SEA-DBS achieves lower PSD and higher reward than the Baseline. For instance, at $75$ steps, SEA-DBS reduces PSD to $302$ and achieves a reward of $23.00\times10^{ -2}$, while the Baseline remains at $328$ PSD and $4.84\times10^{ -2}$ rewards. This indicates that SEA-DBS effectively suppresses abnormal beta activity and maintains higher behavioral performance as the disease progresses while reducing the number of steps required. 

Figure~\ref{fig:testing} compares stimulation at 50 Hz and 30 Hz. At 50 Hz (Figure~\ref{fig:testing_50hz}), stimulation lies above the beta range (13–35 Hz) and more effectively disrupts abnormal oscillations, leading to greater PSD reduction. At 30 Hz (Figure~\ref{fig:testing_30hz}), stimulation overlaps with the pathological beta band and is less effective. These results show that SEA-DBS not only adapts better to changing PD conditions but also selects more effective stimulation strategies. It consistently achieves lower PSD and higher rewards, demonstrating its effectiveness.

\begin{figure}[tb] \centering \includegraphics[width=0.80\linewidth]{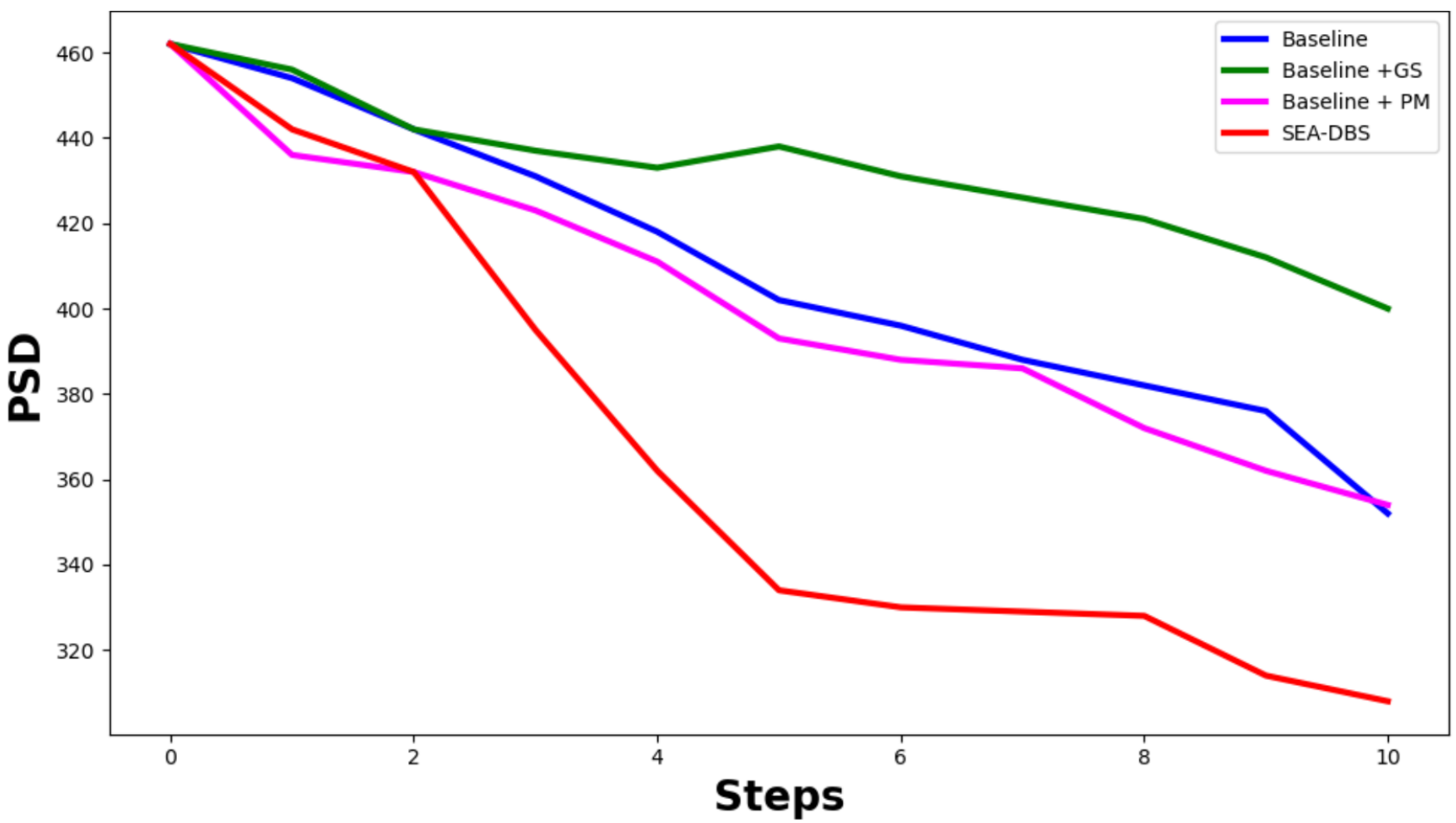} \caption{PSD reduction over 10 stimulation steps for Baseline, Baseline+PM, Baseline+GS, and SEA-DBS. SEA-DBS achieves the most consistent and effective suppression.} \label{fig:ablation} \vspace{-1.5em}  \end{figure}

\noindent \textbf{Ablation Studies.}
Figure~\ref{fig:ablation} compares the performance of four model variants over 10 stimulation steps: Baseline, Baseline+PM (predictive modeling), Baseline+GS (guided sampling), and SEA-DBS (PM+GS).
The Baseline model shows gradual PSD reduction. Adding predictive modeling (baseline+PM) leads to some improvement, though early performance is limited due to noisy predictions and a small training set ($4,500$ samples). Guided sampling (Baseline+GS) helps explore more actions but offers limited gains. SEA-DBS achieves the strongest and most stable reduction. Predictive modeling helps identify promising actions, while guided sampling ensures smoother exploration. Together, they create a more effective learning process that prioritizes long-term benefits and results in better suppression of pathological activity.

\begin{figure}[tb]
\centering \includegraphics[width=0.80\linewidth]{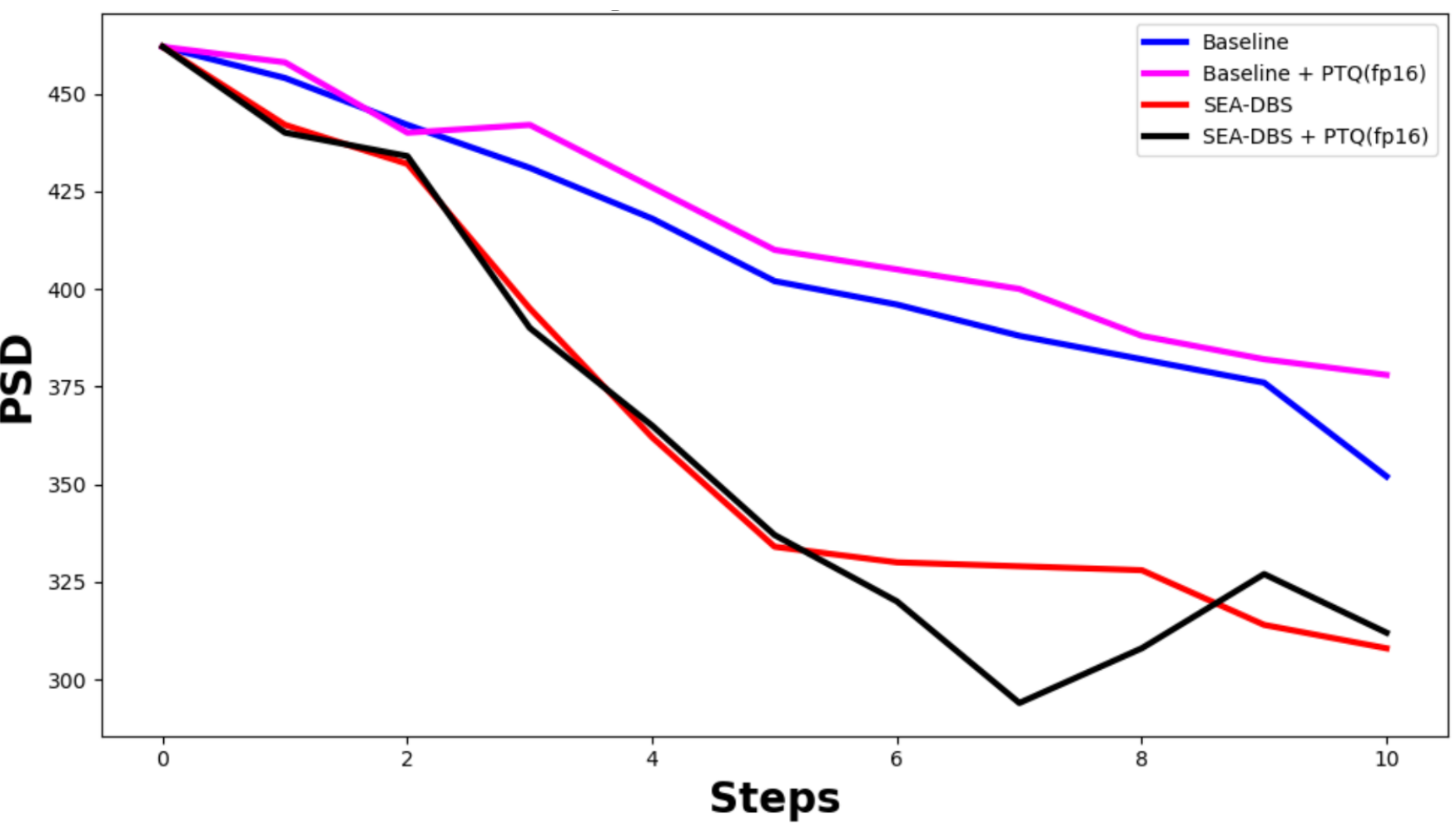} \caption{Performance of SEA-DBS vs Baseline  models after FP16 post-training quantization (PTQ) at 50 Hz.} \label{fig:ptq_fp16} \vspace{-1em} \end{figure}
\noindent \textbf{Model Performance After Quantization.}
Figure~\ref{fig:ptq_fp16} shows the effect of 16-bit floating point (FP16) post-training quantization on SEA-DBS. The quantized model closely matches the full-precision version in reducing PSD over 10 stimulation steps and continues to outperform the Baseline.
Quantization reduces the model size from $65MB$ to $33MB$, enabling deployment on memory-limited devices. SEA-DBS retains its effectiveness after quantization, demonstrating its practicality for real-world, resource-constrained applications.

\section{Conclusion}\label{sec:conclusion}
\noindent This work presented SEA-DBS, a sample-efficient RL-based framework for closed-loop deep brain stimulation in Parkinson’s disease. The proposed system optimized stimulation policies by integrating predictive reward modeling to reduce dependence on real-time feedback and employing Gumbel-Softmax-based exploration for stable learning in binary control spaces. The framework improved personalization by adapting stimulation decisions to individual neural dynamics, enhancing both therapeutic efficacy and energy efficiency. Experiments on a biologically realistic PD model showed that SEA-DBS accelerated convergence, achieved stronger suppression of pathological beta-band activity, and maintained performance under post-training quantization. These findings support the clinical viability of RL-driven neuromodulation in embedded, real-time systems. Future work will aim to further improve model efficiency, validate performance in physical systems, and extend adaptability across diverse patient profiles to advance the translational potential of RL-based aDBS.

\section*{Acknowledgment}
The research reported in this paper was partly supported by the NSF and SC EPSCoR Program under award number (NSF Award \# OIA-2242812 and specific SC EPSCoR grant number 24-GA01 and 25-GA04). The views, perspectives, and content do not necessarily represent the official views of the SC EPSCoR Program or those of the NSF. This work was partially supported by the National Institute of Neurological Disorders and Stroke under Award Number R15NS133859 to Dr. Chunxiu Yu. 
\bibliographystyle{IEEEtran}
\bibliography{main}

\end{document}